# Setting Up the Beam for Human-Centered Service Tasks

Utkarsh Patel, Emre Hatay, Mike D'Arcy, Ghazal Zand, and Pooyan Fazli

Electrical Engineering and Computer Science Department
Cleveland State University, Cleveland, OH 44115
{u.j.patel70,e.hatay,m.m.darcy,g.zand,p.fazli}@csuohio.edu

**Abstract.** We introduce the Beam, a collaborative autonomous mobile service robot, based on SuitableTech's Beam telepresence system. We present a set of enhancements to the telepresence system, including autonomy, human awareness, increased computation and sensing capabilities, and integration with the popular Robot Operating System (ROS) framework. Together, our improvements transform the Beam into a low-cost platform for research on service robots. We examine the Beam on target search and object delivery tasks and demonstrate that the robot achieves a 100% success rate.

## 1  Introduction

In the past decade, there has been a significant growth of research and development on service robots due to their wide range of applications in real life. Although the current working environment of service robots is mostly industrial, these robots are gradually moving from factories and labs to homes, offices, schools, and healthcare facilities. In order for service robots to become an intrinsic part of human environments, they need to be able to perform tasks autonomously and interact with people efficiently.

Autonomy can be challenging to implement on service robots. Unlike many industrial settings, where robots perform repetitive preprogrammed tasks, service robots must act autonomously in dynamic, uncertain, and multi-goal environments. In addition, to be convenient for human users, service robots need to be able to interact with humans in a natural way, by understanding speech and language and reacting to human instructions appropriately.

The Beam is a mobile telepresence system developed by SuitableTech and offers an impressive hardware array for its price. In this paper, we present our modifications to the Beam, which make it possible to use the system as a low-cost platform for research on all aspects of service robotics, including autonomy, human-robot interaction, and multi-robot collaboration and coordination. Our modifications include:

– **Hardware Enhancement:** We add more hardware resources for increased computational and sensing capabilities. These include a laptop to handle expensive computations and multiple depth cameras for increased sensing of the environment.
– **Integration with Robot Operating System (ROS):** We integrate the ROS middleware framework into the Beam to transform it into a programmable research platform.

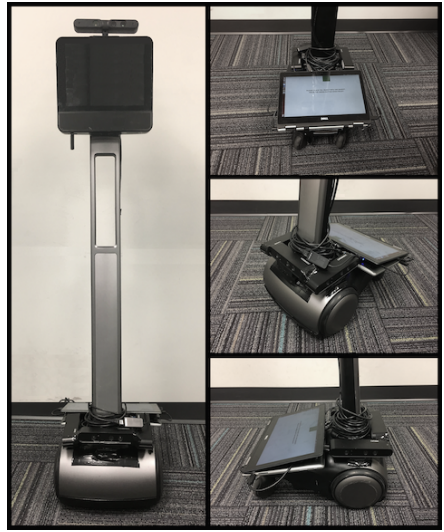

**Fig. 1.** Hardware setup of the Beam

- **Autonomy:** Using ROS, we make the Beam capable of navigating in indoor environments and charging its battery autonomously.
- **Human Awareness and Interaction:** We add human detection, recognition, and tracking capabilities to the Beam, incorporate a speech recognition system to allow the Beam to work with spoken input, and use the laptop's touchscreen to facilitate touch input. We also implement a web interface, shown in Figure 3, and an emailing system to enable remote human users to schedule tasks, monitor, or interact with the robot.

We investigate our improvements to the Beam in target search and object delivery tasks to verify their effectiveness.

## 2   Service Robot Platforms

Unlike industrial robots, which are usually designed to perform routine tasks, recent years have seen various service robots developed to assist human beings in complex, dynamic, and uncertain environments, such as healthcare facilities, hotels, homes, and offices. Nevertheless, these service robots still have limited functionalities. The lack of affordable and commercially available open platforms is a major obstacle in advancing the research on service robotics and has led many robotics researchers and developers to design custom-built platforms with specific capabilities to fit their research agenda. Below, we describe the features of several existing custom-built and commercially available research platforms. Our goal with the Beam is to produce a platform that is easier to set up than the custom platforms but more affordable than the commercial ones.

### 2.1 Custom-built Research Platforms

STanford AI Robot (STAIR) [8] is one of the earliest attempts at building a robot capable of doing service tasks in home and office environments. Research on STAIR was mainly focused on grasping previously unseen objects and opening elevators and new doors.

CoBot [11] is an autonomous service robot capable of performing tasks and interacting with humans robustly in a multi-floor building, and has serviced and traversed more than 1000 km to date. The robot was developed following a novel symbiotic autonomy approach, in which the robot is aware of its perceptual, physical, and reasoning limitations and is able to ask for help from humans proactively.

Herb 2.0 [10] is a robot built to work for and with humans, with a pair of Barrett WAM arms and a mobile base. It is made with a focus on home and office environments, and is therefore built with hardware and software safety features to prevent the robot from taking any unsafe actions. The programming of Herb 2.0 attempts to accomplish the assigned tasks without harming humans, the environment, or itself.

The STRANDS project [4] aims at deploying service robots in security and care scenarios for extended periods. In the security scenario, the robot monitors the environment and people to detect anomalous situations, and in the care scenario, the robot guides visitors and interacts with residents in an elderly care facility. The project focuses on long-term autonomy and learning of service robots in indoor human environments.

BWIBot [6] is a multi-robot platform capable of planning and reasoning in uncertain domains. It also interacts with humans, recognizes human activities, and understands natural language requests and instructions.

To evaluate the performance of domestic service robots, robots can compete in RoboCup@Home [5] challenges in semi-realistic home environments. This competition includes several challenges, each aimed at testing a particular service capability, such as human-robot interaction, manipulation, object recognition, speech recognition, or robust navigation and mapping. KeJia [3], the winner of RoboCup@Home in 2014, is a service robot capable of interacting with humans in natural language and performing manipulation tasks in indoor environments autonomously. This robot was developed following a cognitive approach based on open knowledge available as semistructured data.

### 2.2 Commercial Research Platforms

PR2 [1] is a mobile manipulation platform designed and built by Willow Garage based on Stanford's PR1 robot [14]. It has two 7-DOF arms, an adjustable-height torso, and several sensors, including three cameras and an accelerometer. PR2 is the ROS target platform and has been programmed to do many service tasks, including folding laundry, emptying a dishwasher, and opening doors.

Fetch [13], from Fetch Robotics, is a mobile service robot with a 7-DOF arm, differential drive base, depth camera on its head, and laser range sensor on its base. The robot can carry a 6 kg payload, making it suitable for tasks in warehouses and for a variety of research applications. Fetch is ROS-compatible and has a simulated version compatible with the Gazebo simulation software for easy prototyping.

TIAGo [9] is a mobile manipulator research robot by PAL robotics. In many ways, it is similar to Fetch, with an RGB-D camera, laser range finder sensor, and 7-DOF arm. Unlike Fetch, however, TIAGo defaults to having a five-fingered hand instead of a two-fingered gripper, and it can raise and lower its torso.

SoftBank Robotics' Pepper [7] is a wheeled humanoid robot. It has an omnidirectional base, two 6-DOF arms, and a 3-DOF leg. In addition, Pepper uses an Android tablet on its chest to communicate more effectively and express emotions. Its sensors include six laser sensors, two ultrasound sensors, two tactile sensors in its hands, a 3D camera, and two RGB cameras for human detection and recognition and identifying principal emotions in humans. SoftBank Robotics also makes Romeo [2], a bipedal humanoid robot for assisting elderly people and people with disabilities. Romeo can open doors, climb stairs and reach objects on a table.

## 3  Hardware Setup

The Beam has a 1.34 meter tall polymer body, and weighs about 17.7 kg. It can travel at up to 1 m/s with its a differential wheeled system, in which the front two wheels, powered by motors, can rotate independently and are responsible for steering and moving forward, and the rear two wheels are non-powered swivel casters. The Beam's head consists of a 10-inch screen, four microphones, and a speaker. The head also has two wide-angle 480p HDR cameras: one facing forwards (useful for HRI) and one facing downwards (useful for navigation). Behind the screen is the internal computer, which runs an Intel Celeron 1037U 1.80 GHz dual-core processor and has 1 GB of RAM. The Beam is powered by a 240 Wh battery, which is enough for about two hours of normal teleoperation.

### 3.1  Computation

Because the Beam's internal computer is not powerful enough to run expensive computations in real time, we mounted a laptop on the wheel base to do most of the processing. The laptop has an Intel Core i7-6500U 2.50 GHz quad-core processor and 12 GB of DDR4 2133 MHz RAM.

In our setup, the ROS master node is running on the laptop, so the laptop needs to be able to communicate with the Beam's motor board to send the driving commands. We achieve this by connecting the laptop to the Beam's internal computer through the network and having the internal computer communicate with the motor board. It should be noted that the Beam only has Wifi connectivity exposed by default, so it may struggle in areas of low signal strength and be subject to high latency. We achieved a faster and more robust connection by opening the Beam's head to reveal the internal Ethernet port attached to the motherboard.

### 3.2  Perception

For navigation and localization, we attached three Orbbec Astra Pro cameras to the Beam's base. One faces forward for obstacle detection and localization, and the other two face the side and are used only for localization. For face detection, an Asus Xtion Pro camera is attached on top of the Beam's head, facing the front. All of the cameras are powered by their USB connections, so they can simply be connected to the laptop without a need for an additional power source.

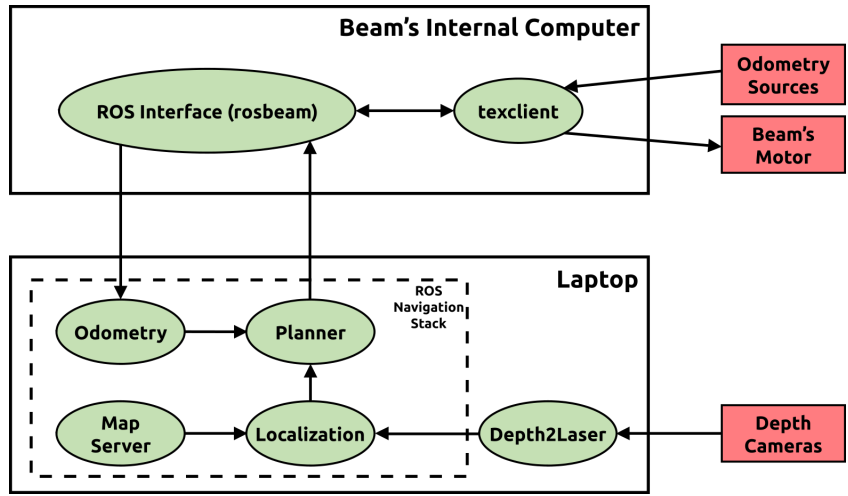

**Fig. 2.** The Beam's navigation and localization module

## 4  Software Setup

We integrate the Beam with ROS, a middleware framework that provides a standardized interface for programming robots. ROS makes it possible to abstract hardware details away, so tasks such as navigation can typically be accomplished just by creating a custom backend to send driving commands to a new hardware platform.

The Beam's operating system is a modified version of Ubuntu, which runs a closed-source program called texclient to communicate with the motor board. Instead of reverse engineering the motor communication protocol from the texclient executable, we used the rosbeam[1] package to send driving commands to the motor board. We made modifications to account for an update to the Beam's OS that broke some features of the package, and the updated source code can be found on our GitHub repository[2]. The rosbeam node injects the driving commands by intercepting the read-write system calls between the texclient process and the serial port device connected to the motor board.

## 5  Autonomy

We add two capabilities to the Beam to establish its autonomy. First, we make it capable of navigating autonomously in indoor environments. Second, we add an autonomous charging routine, so the robot can accomplish long-term missions by charging itself when the battery is low.

---

[1] https://github.com/xlz/rosbeam
[2] https://github.com/people-robots/rosbeam

### 5.1 Autonomous Navigation

To enable autonomous navigation on the Beam, we use the ROS navigation stack[3]. It is a set of algorithms provided by the ROS community that takes in the data from the sensor streams and odometry and outputs safe velocity commands that are sent to the mobile base. The odometry and the sensor streams are used to localize the robot in the map.

The software architecture for localization and navigation is presented in Figure 2. The ROS interface node running on the Beam's computer reads the wheel encoder information from the motor board and publishes them via a ROS topic. The odometry node running on the laptop subscribes to this wheel encoder topic and calculates the odometry, which is later used by the local planner in the core navigation stack. To localize the robot, we use the `amcl`[4] (Adaptive Monte Carlo Localization) package, which takes in the laser scans and the pre-built map and outputs the estimated position of the robot. Because the sensor nodes running on the laptop publish depth images, we convert the images to 2D laser scans using the `depthimage_to_laserscan`[5] package. The 2D map of the environment is constructed using the `rtabmap`[6] package, which uses RGB-D Graph-Based SLAM approach.

Odometry requires tracking both the linear distance traveled and the orientation of the robot in real time. The linear distance is calculated based on the wheel encoders, but the raw encoder values are not reliable because the high inertia of the Beam's long neck causes the front wheels to lift from the ground during navigation. When the front wheels leave the ground, their speed increases for a short period of time, causing the encoders to register distance being traveled while the physical position of the robot does not actually change. To fix this, we defined a threshold for the encoder speed, and if the speed goes above the threshold then we stop increasing the encoder values. To calculate the orientation of the Beam, we used the readings from the built-in accelerometer on the Beam's motor board.

For obstacle detection, we only use the front camera, because running detection with all three cameras uses too much processing power for the laptop to handle well. For localization, however, we use both the front camera and the two side cameras. If only the front camera is used for localization, then the robot does not localize well when navigating through wide corridors. In such situations, the angle of view of the front camera is not large enough to see the side walls and thus the `amcl` node does not get enough data points to correctly localize the robot. The sensor streams from the two cameras facing sideways in opposite directions give `amcl` enough data points to localize the robot in wide corridors.

### 5.2 Autonomous Charging

With a fully charged battery, the Beam can navigate for up to 90 minutes. In order to have long-term autonomy, the Beam must be able to recharge without the need for human intervention. To this end, we implemented a self-charging system using AR

---

[3] http://wiki.ros.org/navigation

[4] http://wiki.ros.org/amcl

[5] http://wiki.ros.org/depthimage_to_laserscan

[6] http://wiki.ros.org/rtabmap

Setting Up the Beam for Human-Centered Service Tasks      7(Augmented Reality) markers. AR markers are paper tags printed with a unique pattern of large black and white squares, making them easy for computer vision algorithms to recognize.

We attach AR markers to each charging base and predefine the coordinates of all the charging stations on the map. The Beam's battery status is monitored using the battery voltage provided by `rosbeam`, and if the battery is running low we set a navigation goal to the nearest charging station. When the robot comes within two meters of the charging station, it rotates to scan for the attached AR marker. As soon as the marker is detected, the robot uses its odometry and the coordinates of the marker to compute the velocity commands to dock itself. To charge the laptop mounted on the back of the robot, we made a circuit to convert and direct the ~12 V potential from the Beam's battery to the ~19 V needed by the laptop's battery, alternating the voltage based on feedback from the laptop.

## 6  Human Awareness and Interaction

We isolated four abilities that we believe the Beam should possess to enable interaction with humans. First, the Beam must be able to be aware of the presence of humans in its environment to know when human-sensitive actions are possible or necessary. It should also have a touchscreen, both to express information to people and to receive tactile inputs. However, limiting communication to touch can be inconvenient and unnatural, so the Beam should also be able to understand speech to enable natural communication with people. Finally, we created a website to allow people to schedule tasks with the robot and monitor its activities even when they are not physically nearby.

### 6.1  Human Detection, Recognition, and Tracking

To detect humans in the environment, we used the `cob_people_detection` package[7], which implements the face detection algorithm proposed by Viola and Jones [12]. The package takes RGB-D image frames as input and outputs information about each detected face, including its pose, bounding box, and optionally a label if face recognition is enabled. We then transform the positions of detected people from the camera coordinate system to positions on the map. This approach allows us to detect, recognize, and track humans in the environment.

### 6.2  Screen and Touchscreen

The laptop we mounted on the Beam has a touchscreen that can be used for sending inputs and displaying information. In addition, we can display information on the Beam's built-in screen, although this is not a touchscreen. Because our current laptop mounting position on the base of the Beam is near the ground, which is not very convenient for users, we are working on adding a touch capability to the built-in screen using a touch overlay.

---

[7] http://wiki.ros.org/cob_people_detection

### 6.3 Speech Recognition and Synthesis

We added speech-to-text capabilities to the Beam to enable it to understand spoken commands. We created a ROS service that takes in audio and outputs plain text. This ROS service is simply a wrapper around our core speech recognition code, which uses the Python `SpeechRecognition` module with the Wit Speech API[8] as the backend to do the speech recognition. In addition to taking speech as input, we enabled the Beam to produce speech as output by installing the `festival` speech synthesis system[9].

### 6.4 Web Interface and Email

Figure 3 shows the Beam's web interface, which allows users to schedule tasks with the Beam and also monitor it through its cameras. The web server is a separate computer in the building, running ROS for communication with the robot and NodeJS to serve the website. When a user selects a task from a drop-down menu on the website along with a set of corresponding task parameters, the NodeJS server receives the request and constructs a ROS message with the task information. This message is sent to the laptop mounted on the Beam, which begins executing the task.

In addition to the website, we also added email capabilities to the robot. This allows it to receive task requests through email, and also allows it to send status notifications when it has completed a task or if it encounters a problem while executing a task. Our email system works through a Gmail account we set up for the robot, using IMAP to receive emails and SMTP to send them.

## 7  Experiments and Results

We extensively examined the Beam's ability to navigate and charge autonomously. In total, the Beam has traversed over 12 km across four buildings connected by bridges at Cleveland State University.

We additionally did proof-of-concept experiments investigating the Beam's performance in two service tasks: target search and object delivery. All experiments took place on the third floor of Fenn Hall Building at Cleveland State University, a map of which is shown in Figure 3.

### 7.1 Target Search

In the target search task, the Beam had to search the building for an object and report its location. To ensure that this would be a test of Beam's abilities and not of the object recognition library implementation, we simplified the object recognition component of this task by attaching AR markers to each object.

For each trial of this task, we hide an AR marker in a random location on the map and instruct the Beam to search for it. The search algorithm consists of the Beam randomly visiting a sequence of predefined possible item locations, illustrated by red stars in the map in Figure 3. Upon reaching a potential location, it rotates for 30 seconds to scan for AR markers. If it sees a marker, it says "I found the object" using its text-to-speech capability.

We conducted 10 trials of this experiment and found that the Beam successfully found the object in all 10.

---

[8] http://wit.ai
[9] http://www.cstr.ed.ac.uk/projects/festival/

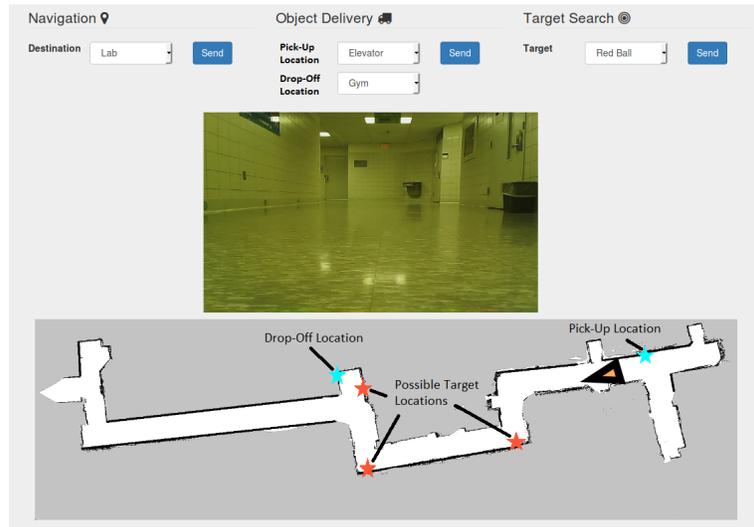

**Fig. 3.** The Beam's web interface. The laser map in the GUI shows the third floor of Fenn Hall building at Cleveland State University

### 7.2 Object Delivery

In the object delivery task, the Beam had to move objects from one location to another. For each object, we gave the Beam a pickup location and a delivery location, each of which were selected randomly from the map. One of the pick-up and drop-off locations are illustrated by cyan stars in the map in Figure 3. We then started the Beam from a random location. The task was for the Beam to travel to the pickup location, get the object, and then travel to the delivery location.

To facilitate the carrying of objects, we attach a small basket to the Beam, and a human is present at the pickup and delivery locations to load and unload the object. When the Beam arrives at the pickup location, it must say "please load the object", after which the human loads the object and presses a button on the laptop touchscreen to indicate that the Beam can move on. Similarly, at the dropoff location the Beam says "please unload the object" and the human uses the touchscreen to indicate when the object had been unloaded.

We tested 10 trials of the object delivery task and recorded the number of times the Beam was able to successfully complete the delivery. We found that the Beam was able to pickup and deliver the object in all 10 trials.

## 8   Conclusion and Future Work

We presented a set of modifications to SuitableTech's Beam telepresence system to transform it into a collaborative autonomous service robot, and made it usable as a low-cost platform for research on different aspects of service robotics, including long-term autonomy and lifelong learning, physical/cognitive/social human-robot interac-tion, multi-robot collaboration and coordination, and human multi-robot interaction. We are currently developing a Universal Robotic Description Format (URDF) model of the Beam to make a simulated version available for the Gazebo simulator software, which would make it much faster and easier to prototype new algorithms and operating environments.